\documentclass{article}

\usepackage{arxiv}

\usepackage[utf8]{inputenc}
\usepackage[T1]{fontenc}
\usepackage{hyperref}
\usepackage{url}
\usepackage{booktabs}
\usepackage{amsmath}
\usepackage{amsfonts}
\usepackage{microtype}
\usepackage{graphicx}
\usepackage{caption}
\usepackage[numbers,sort&compress]{natbib}
\usepackage{doi}
\usepackage{authblk}

\title{CTBakeGS: training-free baking of photorealistic CT volume renderings into 3D Gaussian splats}

\author[1]{Haruki Fukuda\thanks{Corresponding author: \texttt{haruki1995@gunma-u.ac.jp}}}
\author[2]{Constantin Lux}
\author[2]{Wolf Schweitzer}
\author[1]{Tadashi Hosoya}
\author[3]{Lars Ebert}
\affil[1]{Department of Forensic Medicine, Gunma University Graduate School of Medicine, Maebashi, Japan}
\affil[2]{Institute of Forensic Medicine, Universität Zürich, Zürich, Switzerland}
\affil[3]{Biometrics/Forensic Imaging, Zurich Forensic Science Institute, Zurich, Switzerland}

\date{}

\renewcommand{\shorttitle}{CTBakeGS: training-free baking of CT renderings into 3D Gaussian splats}

\hypersetup{
  pdftitle={CTBakeGS: training-free baking of photorealistic CT volume renderings into 3D Gaussian splats},
  pdfauthor={Haruki Fukuda, Constantin Lux, Wolf Schweitzer, Tadashi Hosoya, Lars Ebert},
  pdfkeywords={Computer graphics, Imaging (three-dimensional), Software, Tomography (X-ray computed), Image processing (computer-assisted)},
  colorlinks=true, linkcolor=blue, citecolor=blue, urlcolor=blue
}

\begin{document}
\maketitle

\begin{abstract}
Photorealistic volume rendering of computed tomography (CT) may convey anatomy intuitively, but it needs the CT data, dedicated software and a powerful graphics processor. Three-dimensional Gaussian splatting (3DGS) displays photorealistic scenes in real time on less powerful devices, but existing medical approaches train each scene for minutes to hours. We aimed to convert a CT volume rendering directly into a 3DGS model without training. Splats were placed on an adaptive octree of millimetre-sized cells. Their colour and opacity were computed on the graphics processor with the renderer's lighting, cell opacity was converted analytically into splat opacity, and faint splats were retained stochastically above the 1/255 opacity floor of standard 3DGS rasterisers. Six public CT scans (1.5-mm voxels) were converted with three transfer functions and compared with converged references by peak signal-to-noise ratio (PSNR), structural similarity and a learned perceptual metric. Two-million-splat models reached a PSNR of 32.6 +/- 0.9 dB for bone, 30.1 +/- 1.0 dB for contrast-enhanced vessels and 32.3 +/- 1.5 dB for skin and bone. A conversion pass (usually two or three per model) took 4-30 s on a current laptop and 9-60 s on a desktop with a 4-GB graphics card. On the laptop, the models rendered at 68 frames per second at 800 x 800 pixels. Training-free conversion produced real-time 3DGS models of photorealistic CT renderings in about a minute or less per conversion pass on ordinary computers, with fidelity that we consider sufficient for interactive viewing.
\end{abstract}


\section{Background}

Photorealistic volume rendering of computed tomography (CT) simulates light scattering in tissue and has the capacity to depict anatomy with apparent lifelike depth and shading \cite{kroes2012,jonsson2014,dappa2016,eid2017}. It speeds anatomical understanding compared with conventional CT images and serves surgical planning, teaching, patient communication and documentation \cite{elshafei2019,binder2021,pachowsky2022,ebert2017}. However, it needs the CT data, dedicated software and a powerful graphics processing unit (GPU), so results are usually shared as still images or videos.

Three-dimensional Gaussian splatting (3DGS) represents a scene as millions of semi-transparent Gaussians rasterised in real time \cite{kerbl2023}. Standard 3DGS viewers display such models directly, so a 3DGS model of a CT rendering could be viewed, also in head-mounted displays, without the CT data or renderer. For medical and scientific volumes, most methods optimise Gaussians iteratively against tens to over a thousand rendered views per scene, for minutes to over an hour on a dedicated GPU \cite{niedermayr2024,kleinbeck2025,li2025clipgs,gao2025sixdgs,tang2025,dyken2026,gao2026factorsplat}; one refines an analytic initialisation \cite{li2026vbm}. Feed-forward networks predict Gaussians within seconds but must first be trained on about 100--1,000 volumes with rendered views \cite{gao2026renderfm,wang2026vvgt}. Most optimised models must be rebuilt, or extended \cite{dyken2026,gao2026factorsplat}, when the transfer function changes.

A volume, however, already defines what a Gaussian representation needs: where the material is, how opaque it is and, once lit, its colour. Gaussian kernels have long served direct volume rendering, notably in elliptical weighted average (EWA) splatting \cite{zwicker2001}. Instead of optimising or learning Gaussians from rendered images, we aimed to bake the appearance of a lit CT rendering directly into Gaussians, so that conversion is immediate on ordinary hardware. The contribution is not a new rendering model but the removal of the prerequisites of existing 3DGS pipelines: rendered training views, a dedicated GPU and per-scene training time. We describe CTBakeGS and measure how faithfully it reproduces the original rendering.

\section{Methods}

\subsection{Conversion}

\paragraph{Principle.} A CT volume stores voxel-wise radiodensity values in Hounsfield units. A direct volume rendering is fully defined by three inputs: the CT data, a transfer function that maps Hounsfield units to colour and opacity \cite{max1995}, and the virtual lighting. A 3DGS model, in turn, is a set of Gaussian splats. Each splat is a small, ``soft'', semi-transparent blob in space, described by a position, a size and orientation (scale and rotation; a sphere or a stretched ellipsoid), an opacity and a colour; its opacity is highest at the centre and fades smoothly with distance from it \cite{kerbl2023}. A splat differs from a voxel in three respects. First, a voxel is one sample of a regular grid, so its position and size are fixed by the grid, whereas a splat can be placed anywhere and have any size. Second, a voxel stores a measured quantity, in CT the X-ray attenuation in Hounsfield units, and says nothing about appearance; a splat stores the appearance itself, that is, how that piece of the scene looks. Third, a voxel volume becomes an image only through volume rendering (transfer function, lighting and ray casting) \cite{max1995}, whereas splats are simply projected onto the screen and blended in depth order \cite{kerbl2023,zwicker2001}, which graphics processors do very quickly.

CTBakeGS derives splats directly from the volume: the volume is divided into small cells, each cell becomes one splat, and each splat receives the colour and opacity that the lit volume rendering gives to its cell.

\begin{figure}[t]
  \centering
  \includegraphics[width=1.0\textwidth]{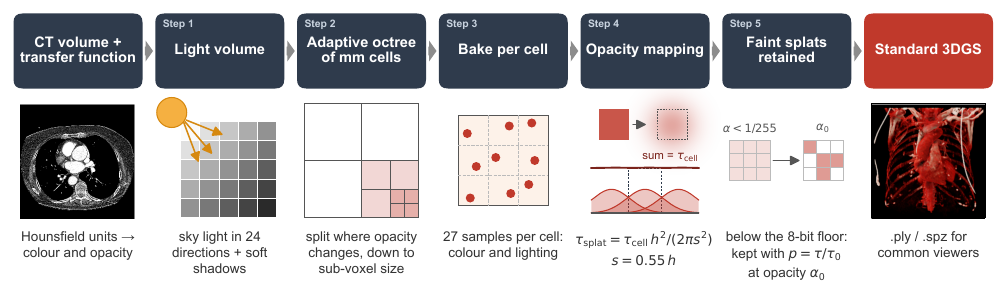}
  \caption{\textbf{Training-free conversion of a CT volume rendering into 3D Gaussian splats.}
  The CT volume is shown with a chosen transfer function, which maps Hounsfield units to colour and opacity; the boxes marked Step~1--5 correspond to the steps in the text. A light volume, shared with the photorealistic renderer, stores the visibility of a sky dome in 24 directions and the transmittance towards a key light with soft shadows. The volume is divided into an adaptive octree of millimetre-sized cubic cells, split only where opacity changes, down to sub-voxel size; the finest size is adjusted to reach the requested number of splats. Each cell is sampled at 27 stratified points with the colour and lighting of the volume renderer, giving one colour with baked lighting and an opacity. Opacity mapping replaces each cell by a Gaussian with the same total opacity, fainter at the centre and wider than the cell ($s = 0.55h$; top); in a profile through a row of cells, the overlapping Gaussians add up to the opacity of the cells (bottom). Faint splats, whose opacity would fall below the 1/255 floor of standard rasterisers, are retained stochastically with raised opacity. The output is a standard 3DGS file (.ply or .spz); the example shows case~2 with the contrast-enhanced vessel transfer function. CT, computed tomography; 3DGS, three-dimensional Gaussian splatting}
  \label{fig:method}
\end{figure}

This takes five steps, described in turn below (Fig.~\ref{fig:method}): (1) virtual lighting is precomputed, so that it can be stored in the splat colours; (2) the volume is divided into cells, which set the positions and sizes of the splats; (3) colour and opacity are sampled within each cell; (4) cell opacity is converted into splat opacity, as neighbouring splats overlap to generate a smooth realistic appearance; and (5) faint splats, which would vanish below the 1/255 opacity floor of standard rasterisers, are retained stochastically with raised opacity. An anti-aliasing compensation is applied when the splats are drawn. The result is written as a standard 3DGS file (.ply or .spz).

\paragraph{Step 1: lighting.} In CTBakeGS, each splat has a single view-independent colour that does not change with the viewing direction, whereas 3DGS models trained from images usually store view-dependent colour. Lighting and shadows must therefore be computed before conversion and stored (baked) in that colour. CTBakeGS uses the lighting of its own photorealistic volume renderer, which approximates global illumination with a precomputed light volume of about 112 cells along the longest axis. For each light-volume cell, rays cast through the transfer-function opacity give the visibility of a sky dome in 24 directions and the transmittance towards a key light, which produces soft shadows. A sample's colour is multiplied by the visibility-weighted sky light plus the transmittance-weighted key light and then tone-mapped. The renderer has a real-time view (one pass per frame), which supplies the colours for conversion, and a progressive view, which averages jittered frames until the image converges.

\paragraph{Step 2: cells.} The cells determine where splats are placed and how large they are. Cells of one voxel each would spend many splats on homogeneous regions and would tie the splat count to the voxel count. Cells are therefore organised as an adaptive octree that is refined only where opacity changes. Starting from cells four voxels wide, a cell is split into eight when its opacity varies strongly, that is, when the maximum transfer-function opacity over its Hounsfield-unit range, $\alpha_{\max}$, exceeds the smaller of the opacities at the two ends of that range by more than $\alpha_{\max}$/2 (and $\alpha_{\max}$ > 0.01). Homogeneous regions thus keep one large splat, whereas boundaries receive small splats, down to sub-voxel size. The split test extends 1.5 cell widths beyond the cell, so that large splats do not protrude through nearby surfaces. Cells are cubes measured in millimetres (unit: the finest voxel spacing), so that thick-slice data do not produce elongated cells.

The number of splats follows from the finest permitted cell size. To reach a chosen target count, this size is adjusted, assuming a power law between cell size and splat count, and the conversion is repeated until the count is within 8\% of the target, usually after two or three conversion passes. If the target falls between two octree levels, as happened once (case~3, bone, one-million target), the count overshoots.

\paragraph{Step 3: colour and opacity.} Each cell is now reduced to one colour and one opacity. The cell is sampled on the GPU at 27 stratified points ($3\times3\times3$). At each point, the transfer function gives the extinction, and the colour is shaded with the lighting model and light volume of the real-time view (Step~1). The point colours are averaged with weights for extinction and for the transmittance towards the most open axis direction, so that points lying deep inside the cell do not darken it.  From the mean extinction $\bar{\sigma}$ and the cell size $h$, the cell opacity is
\begin{equation}
  \alpha_{\mathrm{cell}} = 1 - \exp\!\left(-\bar{\sigma} h\right).
\end{equation}

\paragraph{Step 4: opacity mapping.} Each cell is replaced by an isotropic Gaussian splat with standard deviation s. To avoid gaps and ripples between splats, s is chosen so that each splat extends beyond its cell. Neighbouring splats therefore overlap, and a ray crosses several of them; if every splat simply carried the cell opacity, the model would be too opaque.  With the optical depth of the cell
\begin{equation}
  \tau_{\mathrm{cell}} = -\ln\!\left(1 - \alpha_{\mathrm{cell}}\right),
\end{equation}
the optical depths of a lattice of isotropic Gaussians with spacing h and standard deviation s add up along a ray to that of the volume when each splat carries
\begin{equation}
  \tau_{\mathrm{splat}} = \tau_{\mathrm{cell}}\,\frac{h^{2}}{2\pi s^{2}}.
\end{equation}
With $s = 0.55h$, the narrowest width for which Gaussians on a lattice of spacing h sum to a constant optical depth within 1\% along each lattice axis, the splat opacity is
\begin{equation}
  \alpha_{\mathrm{splat}} = 1 - \exp\!\left(-0.526\,\tau_{\mathrm{cell}}\right).
\end{equation}

\paragraph{Step 5: faint splats.} Standard 3DGS rasterisers skip splats whose opacity is below 1/255, and compact formats such as .spz store opacity with 8 bits, so a splat with $\alpha_{\mathrm{splat}}$ below this floor vanishes. In a thin translucent layer, every cell is below the floor, although many such cells along an oblique ray add up to a visible veil; discarding them removes the whole layer. CTBakeGS therefore keeps a faint splat ($\alpha_{\mathrm{splat}}$ < $\alpha_0$, with $\alpha_0$ = 0.01, about 2.5/255, so that it stays above the floor after the anti-aliasing compensation) with probability $\tau_{\mathrm{splat}}/\tau_0$, where $\tau_0 = -\ln(1-\alpha_0)$, and assigns it the opacity $\alpha_0$. The expected optical depth of the lattice is unchanged, and fewer splats are needed. The choice uses a hash of the cell index, so the conversion remains deterministic.

\paragraph{Anti-aliasing compensation.} To avoid aliasing, 3DGS rasterisers enlarge each projected Gaussian by a small screen-space filter (variance 0.3 pixel$^2$). This makes small splats appear thicker than intended \cite{yu2024mip}. As in Mip-Splatting \cite{yu2024mip}, the splat opacity is therefore multiplied at drawing time by
\begin{equation}
  w = \sqrt{\frac{\det(\Sigma)}{\det(\Sigma + 0.3I)}},
\end{equation}
where $\Sigma$ is the projected covariance and I the identity matrix.

\subsection{Implementation}

CTBakeGS is a single HTML file (JavaScript, WebGL 2) that runs locally in a web browser. It is available at \url{https://github.com/BocchiSlice/CTBakeGS}; version 1.0 was used for all measurements.

\subsection{Evaluation}

\paragraph{Data.} We used six CT scans from two scanner vendors in the TotalSegmentator dataset v2 (CC BY 4.0) \cite{wasserthal2023}: case~1 (s0289), head CT angiography; case~2 (s0138), heart and thoracic aorta; case~3 (s1340), thorax to pelvis; case~4 (s0970), polytrauma; case~5 (s1121), pelvis-to-leg CT angiography; case~6 (s0931), neck-to-leg CT angiography (Supplementary Table~\ref{tab:cases}). All have 1.5-mm isotropic voxels (215--519 slices; 4.9--56.7 million voxels). For the thick-slice test, the 1.5-mm slices of case~1 were also averaged along the slice direction in groups spanning 5~mm, giving voxels of $1.5\times1.5\times5$~mm.

\paragraph{Views and reference.} Each scan was rendered with three preset transfer functions (bone; contrast-enhanced vessels; skin and bone), giving 18 scenes plus three for the 5-mm data. For each scene, 48 wide views (16 azimuths at three elevations) and 16 close-up views were rendered at $800\times800$ pixels on a black background. Close-up views were centred on visible surface points chosen by farthest-point sampling, at 0.3 times the wide distance. The reference image for each view was rendered by the progressive view of the renderer (Step~1), from the same volume, transfer function and lighting, by averaging 128 jittered frames. The same view averaged over 256 frames differed from it by a peak signal-to-noise ratio (PSNR; see Metrics) of 54.7--58.7~dB, a root-mean-square (RMS) difference below half an 8-bit grey level and far smaller than the differences of the 3DGS models, so the reference was considered converged.

The 3DGS models were rasterised from the same cameras.

\paragraph{Metrics.} Each 3DGS model was compared with the reference by PSNR, computed over the whole image and over the foreground (pixels non-black in either image), the structural similarity index (SSIM) \cite{wang2004} and the learned perceptual image patch similarity (LPIPS, VGG backbone) \cite{zhang2018}. PSNR is 20 $\log_{10}$ of the full-scale value divided by the RMS pixel difference; 20~dB more means a tenfold smaller RMS difference, and 32~dB corresponds to about 2.5\% of full scale, or 6 of 255 grey levels in an 8-bit image. We also recorded splat counts, conversion times and frame rates; frame rates were measured in the built-in WebGL rasteriser of CTBakeGS.

\paragraph{Variants.} To test how splat count and the individual conversion steps affect fidelity, we evaluated models with about 1, 2 (main setting) and 3 million splats, and five ablations, each of which removes or replaces one step:
\begin{itemize}
  \item no opacity mapping (Step~4; cell opacity used as is);
  \item no anti-aliasing compensation;
  \item without faint splats (Step~5; splats below the 8-bit floor discarded);
  \item a uniform grid with one cell per voxel instead of the adaptive octree (Step~2), compared with the octree at the same number of splats;
  \item voxel-unit instead of millimetre cells (Step~2; 5-mm data only).
\end{itemize}

\paragraph{Hardware.} Measurements used Chrome 155 on a 14-inch Apple MacBook Pro (2024; M4 Max, 14-core CPU, 32-core GPU, 36~GB memory, macOS 26.6). Speed was also measured on a Windows desktop (Chrome 155; 2018; Intel Core i7-8700, 16~GB memory, NVIDIA GeForce GTX 1050 Ti, 4~GB, Windows 11) and conversion time on a 13-inch Apple MacBook Air (Chrome 137; 2017; 1.8-GHz dual-core Intel Core i5, Intel HD Graphics 6000, 8~GB memory, macOS 12.6).

\paragraph{Statistics.} Metrics were averaged over the views of each scene, and the case was the unit of analysis (mean$\pm$standard deviation over six cases). Variants were compared with the main setting by the two-sided exact Wilcoxon signed-rank test; with six cases the smallest attainable p value is 0.031, and p values are exploratory and not adjusted for multiple comparisons. Analyses used Python 3.9 with NumPy 2.0, SciPy 1.13, pandas 2.3 and lpips 0.1.4 (PyTorch 2.8).

\section{Results}

\subsection{Fidelity}

\begin{figure}[p]
  \centering
  \includegraphics[width=0.86\textwidth]{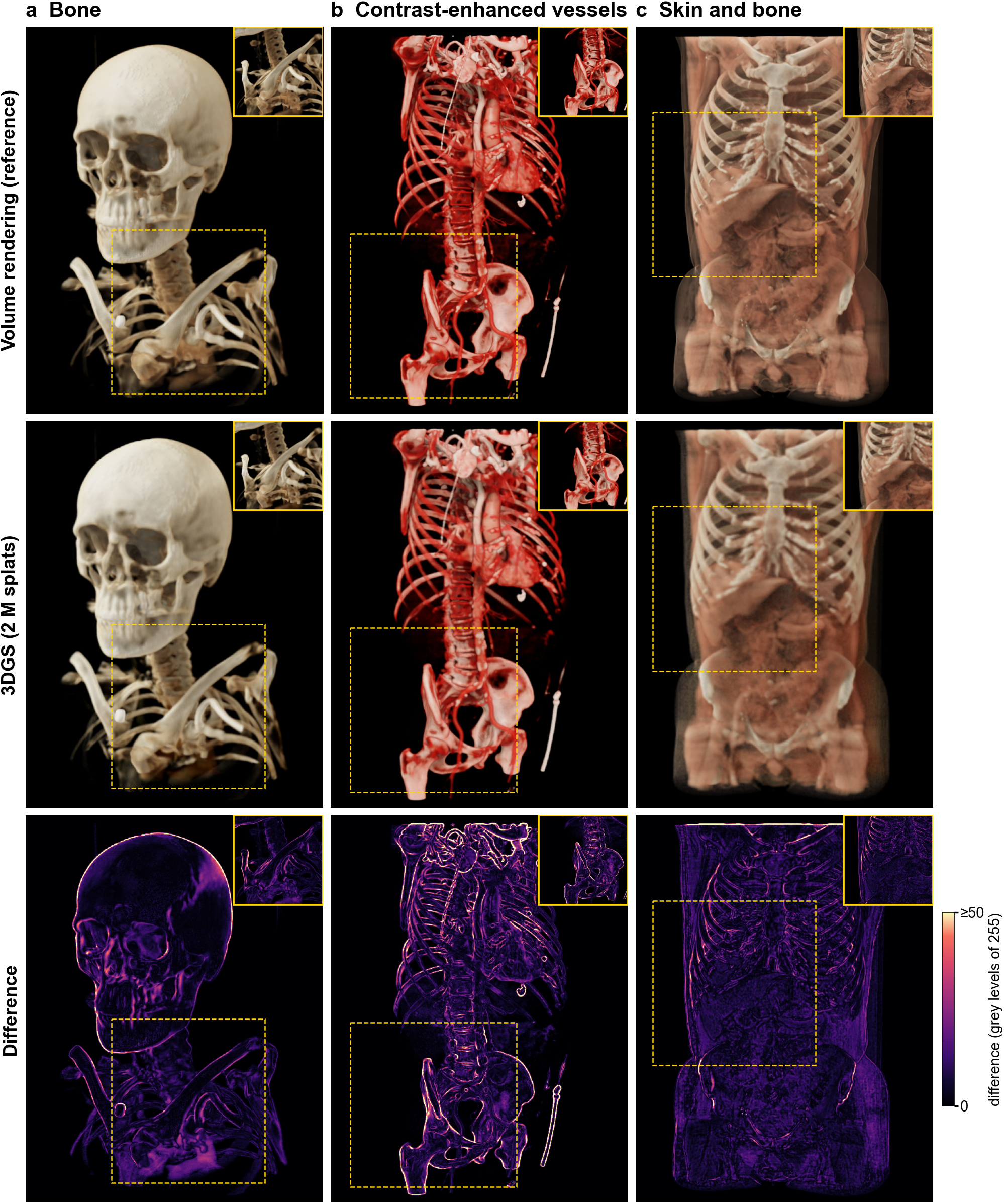}
  \caption{\textbf{Representative results: converged references, 3DGS models and their differences.}
  (a) Bone, case~1; (b) contrast-enhanced vessels, case~6; (c) skin and bone, case~3; one wide view each, cropped to the body. Top: converged photorealistic volume rendering (reference); middle: 3DGS model with two million splats; bottom: difference from the reference (mean absolute difference over the colour channels, 0 to $\geq$ 50 grey levels of 255). Insets (yellow frames) show a close-up view of the same scene; dashed boxes mark the region it covers in the wide view. Errors lie mainly along the edges of dense structures for bone and vessels and are spread over the translucent layer for skin and bone. 3DGS, three-dimensional Gaussian splatting}
  \label{fig:results}
\end{figure}

Splat counts were within 8\% of the targets in every scene except one (case~3, bone, 1.37 million for the one-million target). With two million splats, wide-view PSNR was 32.6$\pm$0.9~dB for bone, 30.1$\pm$1.0~dB for contrast-enhanced vessels and 32.3$\pm$1.5~dB for skin and bone, with SSIM of 0.960--0.980 and LPIPS of 0.028--0.091 (Table~\ref{tab:quality}; Fig.~\ref{fig:results}). Close-up views were about 3~dB lower; the real-time photorealistic view reached 42--47~dB. Errors lay mainly along the edges of dense structures for bone and vessels and were spread over the translucent layer for skin and bone (Fig.~\ref{fig:results}). Fidelity was lowest in close-up views of case~4, the largest volume, for bone and vessels (27.1 and 24.0~dB), and in wide views of case~2 for skin and bone (29.8~dB; other cases 32.0--34.5~dB; Table~\ref{tab:percase}).

\begin{table}[t]
  \centering
  \caption{\textbf{Image quality of the 3DGS models against converged photorealistic references}}
  \label{tab:quality}
  \small
  \setlength{\tabcolsep}{4pt}
  \begin{tabular}{lcccccc}
    \toprule
    Look, views & \multicolumn{3}{c}{PSNR (dB)} & SSIM & LPIPS & PSNR, real-time \\
    \cmidrule(lr){2-4}
     & 1\,M & 2\,M & 3\,M & 2\,M & 2\,M & photorealistic (dB) \\
    \midrule
    Bone, wide & 32.02$\pm$1.42 & 32.61$\pm$0.89 & 32.52$\pm$0.86 & 0.980$\pm$0.005 & 0.028$\pm$0.014 & 44.61$\pm$1.26 \\
    Bone, close-up & 28.46$\pm$1.87 & 29.63$\pm$1.55 & 30.21$\pm$1.42 & 0.949$\pm$0.012 & 0.135$\pm$0.041 & 45.89$\pm$1.64 \\
    Vessels, wide & 29.17$\pm$1.15 & 30.09$\pm$0.98 & 30.43$\pm$0.87 & 0.960$\pm$0.011 & 0.036$\pm$0.016 & 42.28$\pm$1.47 \\
    Vessels, close-up & 25.23$\pm$1.55 & 26.56$\pm$1.62 & 27.29$\pm$1.60 & 0.895$\pm$0.023 & 0.150$\pm$0.019 & 43.89$\pm$1.51 \\
    Skin and bone, wide & 31.78$\pm$1.80 & 32.32$\pm$1.51 & 32.19$\pm$1.18 & 0.961$\pm$0.017 & 0.091$\pm$0.040 & 47.12$\pm$0.79 \\
    Skin and bone, close-up & 28.24$\pm$1.04 & 29.09$\pm$1.01 & 29.43$\pm$0.94 & 0.925$\pm$0.013 & 0.314$\pm$0.029 & 46.70$\pm$1.19 \\
    \bottomrule
  \end{tabular}

  \vspace{4pt}
  \begin{minipage}{\textwidth}
  \footnotesize Mean\,$\pm$\,standard deviation over six cases; metrics were first averaged over the 48 wide or 16 close-up views of each scene. Models had approximately 1, 2 or 3 million (M) splats. Higher PSNR and SSIM and lower LPIPS indicate closer pixel-level agreement with the reference; they do not measure anatomical accuracy. LPIPS, learned perceptual image patch similarity; PSNR, peak signal-to-noise ratio; SSIM, structural similarity index; 3DGS, three-dimensional Gaussian splatting.
  \end{minipage}
\end{table}

\begin{table}[t]
  \centering
  \caption{\textbf{Per-case image quality with two million splats, PSNR (dB) / SSIM / LPIPS}}
  \label{tab:percase}
  \scriptsize
  \setlength{\tabcolsep}{3pt}
  \begin{tabular}{llcccccc}
    \toprule
    Look & Views & Case 1 & Case 2 & Case 3 & Case 4 & Case 5 & Case 6 \\
    \midrule
    Bone & wide & 33.6/0.982/0.031 & 32.0/0.971/0.054 & 33.1/0.980/0.023 & 31.1/0.979/0.022 & 33.1/0.980/0.018 & 32.6/0.985/0.017 \\
    Bone & close-up & 30.0/0.961/0.178 & 31.2/0.948/0.192 & 30.0/0.945/0.124 & 27.1/0.935/0.118 & 30.9/0.965/0.085 & 28.5/0.941/0.114 \\
    Vessels & wide & 30.2/0.966/0.042 & 30.4/0.942/0.065 & 30.3/0.955/0.034 & 28.3/0.962/0.030 & 31.3/0.969/0.019 & 30.1/0.969/0.024 \\
    Vessels & close-up & 27.4/0.925/0.156 & 28.7/0.911/0.174 & 25.9/0.861/0.163 & 24.0/0.877/0.147 & 27.4/0.903/0.120 & 26.1/0.893/0.138 \\
    Skin and bone & wide & 32.3/0.968/0.088 & 29.8/0.927/0.170 & 32.9/0.963/0.087 & 32.0/0.968/0.070 & 34.5/0.976/0.064 & 32.5/0.967/0.069 \\
    Skin and bone & close-up & 29.4/0.942/0.306 & 28.7/0.935/0.356 & 29.9/0.922/0.322 & 28.2/0.904/0.314 & 30.5/0.926/0.267 & 27.9/0.921/0.316 \\
    \bottomrule
  \end{tabular}

  \vspace{4pt}
  \begin{minipage}{\textwidth}
  \footnotesize Metrics were averaged over the 48 wide or 16 close-up views of each scene; case numbers as in Supplementary Table~\ref{tab:cases}. LPIPS, learned perceptual image patch similarity; PSNR, peak signal-to-noise ratio; SSIM, structural similarity index.
  \end{minipage}
\end{table}

\subsection{Splat count and speed}

Compared with two million splats, one million lowered close-up PSNR in all cases for bone (median $-$1.2~dB), vessels ($-$1.3~dB) and skin and bone ($-$0.9~dB), and three million raised it in all cases (+0.6, +0.8 and +0.4~dB); close-up LPIPS followed the same pattern (Supplementary Table~\ref{tab:paired}). A two-million-splat pass took 14.3~s on average (4.2--29.5~s), and rendering ran at 68 frames per second ($800\times800$ pixels; Table~\ref{tab:speed}). A desktop with a 4-GB graphics card needed 29.4~s per pass and rendered 28 frames per second. The 2017 laptop needed 106~s (30--216~s) and rendered 7 frames per second.

\begin{table}[t]
  \centering
  \caption{\textbf{Splat counts, conversion times and frame rates}}
  \label{tab:speed}
  \small
  \begin{tabular}{lccc}
    \toprule
    Model & 1\,M & 2\,M & 3\,M \\
    \midrule
    Splats (million) & 1.04 (0.92--1.37) & 2.00 (1.89--2.15) & 2.97 (2.76--3.23) \\
    Conversion time, MacBook Pro (s) & 11.0 (2.2--24.0) & 14.3 (4.2--29.5) & 17.1 (3.1--30.7) \\
    Frame rate, MacBook Pro (fps) & 122 (97--137) & 68 (62--74) & 48 (43--53) \\
    Conversion time, Windows desktop (s) & 23.1 (4.6--50.0) & 29.4 (8.8--60.2) & 35.1 (7.4--63.0) \\
    Frame rate, Windows desktop (fps) & 52 (36--60) & 28 (22--33) & 19 (16--22) \\
    Conversion time, MacBook Air 2017 (s) & 85 (22--193) & 106 (30--216) & ---$^{\mathrm{a}}$ \\
    \bottomrule
  \end{tabular}

  \vspace{4pt}
  \begin{minipage}{0.9\textwidth}
  \footnotesize Mean (range) over six cases $\times$ three looks. Conversion time is one conversion pass at the final cell size; reaching the target splat count usually took two or three passes; frame rates at $800\times800$ pixels in Chrome. $^{\mathrm{a}}$Not reported: three-million-splat conversions completed in only 10 of 18 scenes on this computer, the others exceeding its graphics memory. The .ply files hold 68 bytes per splat (128--146~MB for the 2-M models).
  \end{minipage}
\end{table}

\subsection{Ablations}

\begin{figure}[p]
  \centering
  \includegraphics[width=0.9\textwidth]{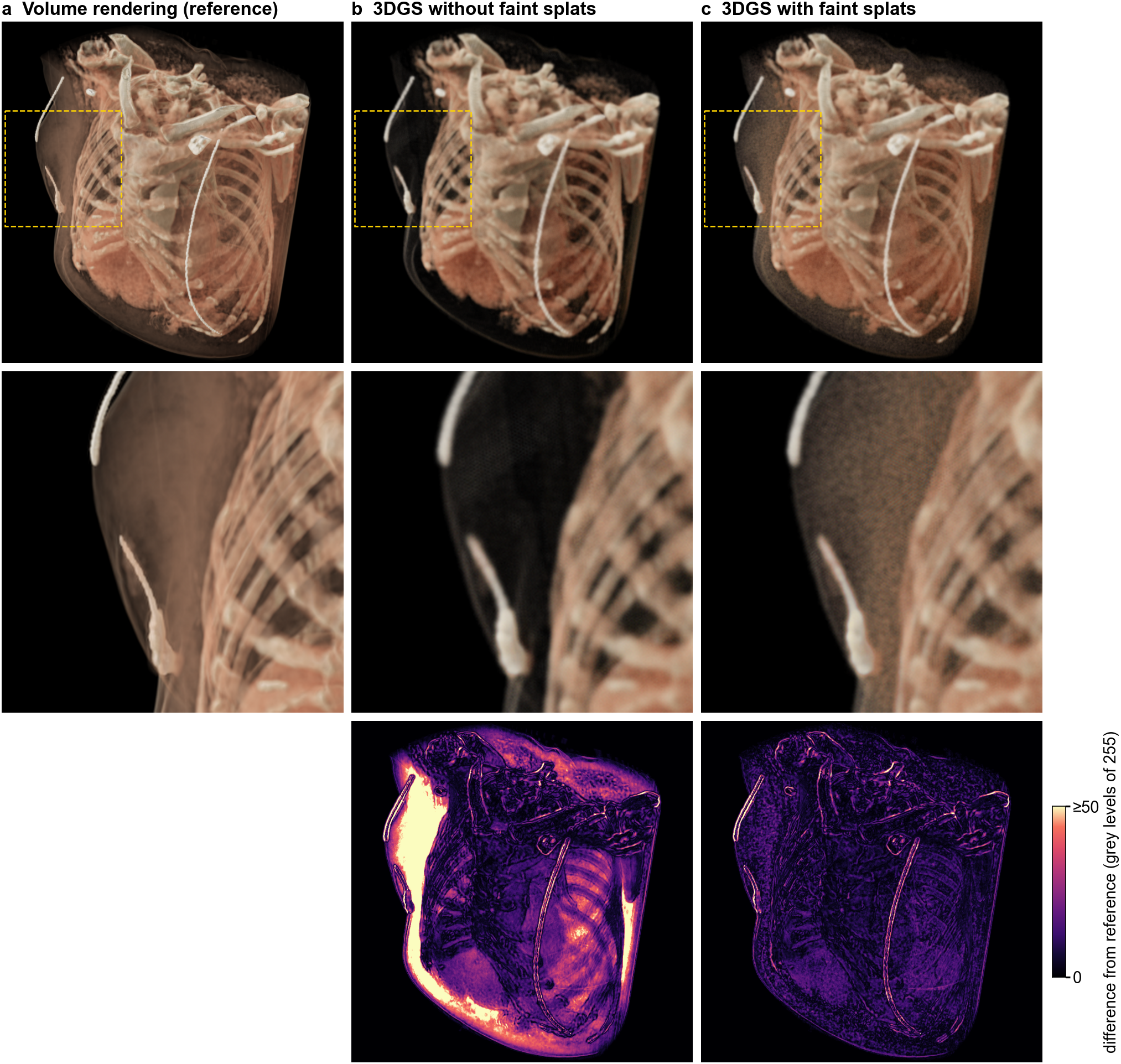}
  \caption{\textbf{Faint splats: case 2, skin and bone, two million splats.}
  (a) Converged photorealistic volume rendering (reference); (b) 3DGS model without faint splats (ablation of Step~5); (c) 3DGS model with faint splats (CTBakeGS). Top row: one wide view, cropped to the body; middle row: close-up of the dashed box; bottom row: difference from the reference as in Fig.~\ref{fig:results} (0 to $\geq$ 50 grey levels of 255). Without them, the thin outer soft-tissue layer, whose splats fall below the 8-bit opacity floor, is missing and the ribs show through (PSNR 22.5~dB in this view); with them it is restored (30.1~dB), with a fine grain visible at close range. 3DGS, three-dimensional Gaussian splatting; PSNR, peak signal-to-noise ratio}
  \label{fig:faint}
\end{figure}

Without opacity mapping, PSNR was lower in all cases for every look (median $-$4.5 and $-$5.1~dB for bone in wide and close-up views, $-$1.3 and $-$1.9~dB for vessels, $-$5.3 and $-$5.9~dB for skin and bone; Supplementary Table~\ref{tab:paired}). Without anti-aliasing compensation, wide-view PSNR was lower in all cases for bone ($-$3.8~dB) and vessels ($-$0.9~dB), with little change for skin and bone and in close-up views. At equal splat count, a uniform grid with one cell per voxel was as faithful as the octree for vessels and skin and slightly better for bone (+0.4 and +0.6~dB, all cases). The voxel grid, however, fixes the splat count (0.14--2.86 million); in the two smallest volumes, two-million-splat octree models exceeded it by 1.3--2.7~dB for bone and vessels. With 5-mm slices, millimetre cells gave PSNR similar to the 1.5-mm data (bone, wide views: 34.3 vs 33.6~dB), whereas voxel-unit cells gave 22.5~dB (Supplementary Table~\ref{tab:thick}; per-case details in Supplementary Table~\ref{tab:cases}). Without faint splats, PSNR was lower in all cases in wide views (bone $-$0.3~dB, vessels $-$0.4~dB, skin and bone $-$1.6~dB) and in close-up views of vessels and skin and bone ($-$0.5 and $-$1.4~dB), with little change in LPIPS; the loss was largest in case~2, where wide-view PSNR for skin and bone fell from 29.8 to 23.8~dB (Table~\ref{tab:faint}; Fig.~\ref{fig:faint}).

\begin{table}[t]
  \centering
  \caption{\textbf{Per-case PSNR (dB) with two million splats, with / without faint splats}}
  \label{tab:faint}
  \small
  \setlength{\tabcolsep}{4pt}
  \begin{tabular}{llcccccc}
    \toprule
    Look & Views & Case 1 & Case 2 & Case 3 & Case 4 & Case 5 & Case 6 \\
    \midrule
    Bone & wide & 33.6 / 33.4 & 32.0 / 31.6 & 33.1 / 32.5 & 31.2 / 30.8 & 33.1 / 32.9 & 32.6 / 32.6 \\
    Bone & close-up & 30.0 / 29.7 & 31.2 / 30.7 & 30.0 / 29.2 & 27.1 / 26.7 & 30.9 / 30.0 & 28.5 / 28.5 \\
    Vessels & wide & 30.2 / 30.1 & 30.4 / 29.7 & 30.3 / 29.9 & 28.3 / 27.9 & 31.3 / 30.9 & 30.1 / 29.7 \\
    Vessels & close-up & 27.4 / 27.3 & 28.7 / 27.7 & 25.9 / 25.4 & 24.0 / 23.5 & 27.4 / 26.6 & 26.1 / 25.6 \\
    Skin and bone & wide & 32.3 / 29.8 & 29.8 / 23.8 & 32.9 / 31.4 & 32.0 / 31.2 & 34.5 / 33.7 & 32.5 / 30.7 \\
    Skin and bone & close-up & 29.4 / 27.4 & 28.7 / 23.7 & 29.9 / 28.6 & 28.2 / 27.2 & 30.5 / 29.8 & 27.9 / 26.5 \\
    \bottomrule
  \end{tabular}

  \vspace{4pt}
  \begin{minipage}{\textwidth}
  \footnotesize PSNR averaged over the 48 wide or 16 close-up views of each scene; case numbers as in Supplementary Table~\ref{tab:cases}. With: the full method (Step~5); without: splats below the 8-bit opacity floor discarded (ablation). PSNR, peak signal-to-noise ratio.
  \end{minipage}
\end{table}

\section{Discussion}

CTBakeGS converted CT volume renderings into standard 3DGS models in about a minute or less per conversion pass, usually two or three passes per model, on a current laptop and on a 2018 desktop with a 4-GB graphics card, without training or per-scene optimisation. The idea is simple, but three steps made it work: the opacity mapping for overlapping splats, the baked lighting and the retention of faint splats above the 8-bit opacity floor. It is deterministic as every splat derives from its cell's voxels; a new transfer function needs only another conversion; and, with the virtual lighting baked in, standard viewers show the lit result. Because the viewer runs in a web browser, no software has to be installed, and a converted model could be packaged with the viewer page and sent, for example, to patients as their CT scan in a form that runs on ordinary devices. Because splats are a common format, a volume rendering could also be combined with splat models of other origin, such as reconstructions of a crime scene.

The fidelity results may define the scope of use. Bone reached about 33~dB in wide and 30~dB in close-up views, corresponding to an RMS pixel difference of about 2.3\% and 3.3\% of full scale, or 6 and 8 of 255 grey levels, and skin and bone about 29~dB in close-up views (3.5\%; 9 of 255), whereas the real-time volume renderer reached 42--47~dB (0.45--0.8\%; 1--2 of 255). Contrast-enhanced vessels were weakest in close-up views (27~dB; 4.7\%; 12 of 255), where the edges of dense bone and thin vessels were smoothed. In case~4, the largest volume, the same splat budget gave coarser cells, and close-up fidelity rose with more splats (vessels 22.8, 24.0 and 24.8~dB with 1, 2 and 3 million splats), so large volumes may need more splats.

Values reported for trained models (about 21--37~dB \cite{niedermayr2024,li2025clipgs,gao2026renderfm}) used different data, resolutions and references and are not directly comparable. Optimisation \cite{niedermayr2024,kleinbeck2025,li2025clipgs,gao2025sixdgs,tang2025,dyken2026,gao2026factorsplat,li2026vbm}, feed-forward networks \cite{gao2026renderfm,wang2026vvgt} and hybrid streaming \cite{kleinbeck2026} remain preferable when the highest fidelity matters and their cost is acceptable. CTBakeGS does not compete with them on fidelity but removes their prerequisites.

Faint splats were the main source of error in translucent layers. Standard 3DGS rasterisers discard opacities below 1/255, and the thin outer soft-tissue layer of the skin-and-bone look consists of such faint cells, which are visible in the volume rendering only because many of them add up along oblique rays. Without the retention of Step~5, this layer was largely missing in case~2 and the ribs showed through it (Fig.~\ref{fig:faint}); its wide-view PSNR fell from 29.8 to 23.8~dB, and bone and vessels in that case lost 0.5 and 0.7~dB (Table~\ref{tab:faint}). In exploratory runs, lowering the discard threshold to 0.001 or not splitting faint cells did not restore the layer (+0.1 and +0.3~dB); retaining faint splats stochastically with raised opacity did. The retained splats are individually denser than the tissue they represent, which gives the layer a fine grain at close range. Isotropic splats may still represent thin translucent sheets poorly, and 3DGS rasterisers sort whole splats rather than samples along each ray \cite{radl2024}. The translucent skin layer is itself a simplification: real skin has a complex, spectrally varying reflectance with subsurface scattering in its upper layers \cite{gitlina2020}, and easy-to-use, realistic skin rendering from CT remains for future research.

At equal splat count, the octree was not more faithful than a uniform grid. Its benefit is that the splat count can be chosen, for example two million for every volume, instead of being dictated by the number of voxels, which ranged from 0.14 to 2.86 million here.

This study has limitations. Six cases from one public dataset were used, with 1.5-mm voxels, which are slightly coarser than the maximum slice thickness of 1.25~mm proposed for three-dimensional reconstruction in forensic casework \cite{ford2016}; clinical thin-slice scans have up to ten times more voxels and convert more slowly. We also stopped at three million splats; close-up fidelity was still rising at that count, so more splats may improve it further at the cost of file size and frame rate, whereas wide-view fidelity had levelled off. The evaluation has two boundaries of its own. Fidelity was measured against the converged view of the same renderer, whose lighting only approximates global illumination, so agreement with other photorealistic renderers is unknown, and PSNR, SSIM and LPIPS do not establish anatomical accuracy. Nor did we test the use of the models: whether they aid teaching or communication, or are suitable for diagnostic reading, requires reader studies. Finally, two choices in the method were made rather than optimised. The opacity floor of the retained faint splats ($\alpha_0$ = 0.01) is roughly the smallest value that stays above the 1/255 cut-off after the anti-aliasing compensation, which reduces the drawn opacity of small splats, and the two interact; and lighting, transfer function and clipping are fixed at conversion, so that view-independent colours lose specular highlights.

Several of these limitations point to the next steps. A larger series, including clinical thin-slice scans, should be tested, and conversion of thin-slice and very large volumes should become faster, for example by caching the Hounsfield-unit ranges of the volume for the octree. It also remains to be determined which computer is the minimum for displaying (rather than converting) the models, how they compare with volume rendering and path tracing in this respect and in compressibility. Whether fine structures such as hairline fractures are preserved also needs to be tested, and the error between the model and the rendering could be used to place more splats where details are lost. Fidelity could rise further with view-dependent colours (spherical harmonics), which would restore highlights, and with flat or anisotropic splats for thin translucent layers \cite{huang2024}. The converted model could also be edited directly, cut with a clip plane or recoloured by segmented organ, without reconverting. Above all, reader studies are needed to show whether the models aid teaching and communication.

\section{Conclusions}

We could demonstrate that photorealistic CT volume renderings can be converted directly into standard 3D Gaussian splats, without training, and displayed in a web browser. Conversion took about a minute or less per pass on a current laptop and on a 2018 desktop with a 4-GB graphics card, and under 4 minutes per pass on a 2017 laptop with integrated graphics. The models reproduce the rendering with root-mean-square errors of about 2.3--3.1\% of full scale in wide views. This approach could make complex CT data easier to share and view interactively without the CT data or a volume renderer, with potential value for communication and teaching.

\section*{Abbreviations}
3DGS, three-dimensional Gaussian splatting; CT, computed tomography; EWA, elliptical weighted average; GPU, graphics processing unit; LPIPS, learned perceptual image patch similarity; PSNR, peak signal-to-noise ratio; RMS, root-mean-square; SSIM, structural similarity index.

\section*{Acknowledgements}
Not applicable.

\section*{Declaration of the use of artificial intelligence}
The software code was written with the assistance of an artificial intelligence model (Claude Opus 5.5, Anthropic), which was also used to help draft, edit and translate the manuscript. The authors reviewed, tested and revised all code and text and take full responsibility for the content of this article.

\section*{Authors' contributions}
Conceptualization, H.F., W.S., C.L.; methodology, H.F.; software, H.F. and W.S. ; validation, H.F.; formal analysis, H.F.; investigation, H.F.; resources, H.F.; data curation, H.F.; writing---original draft preparation, H.F.; writing---review and editing, H.F., C.L., W.S., T.H. and L.E. ; visualization, H.F.; supervision, H.F.; project administration, H.F.; funding acquisition, H.F. All authors have read and agreed to the published version of the manuscript.

\section*{Funding}
This work was supported in part by KAKENHI Grant Number 25K20648 and Overseas Research Fellowships (grant number 202560103) from the Japan Society for the Promotion of Science.

\section*{Availability of data and materials}
The source code and the evaluation scripts are available at \url{https://github.com/BocchiSlice/CTBakeGS}. The evaluation outputs are available from the corresponding author on reasonable request.

\section*{Declarations}

\paragraph{Ethics approval and consent to participate.} Not applicable. This study used only publicly available, de-identified CT data (TotalSegmentator dataset, CC BY 4.0).

\paragraph{Consent for publication.} Not applicable.

\paragraph{Competing interests.} The authors declare that they have no competing interests.

%% ---------------------------------------------------------------
\clearpage
\appendix
\renewcommand{\thetable}{S\arabic{table}}
\setcounter{table}{0}
\section*{Supplementary material}

\begin{table}[h]
  \centering
  \caption{\textbf{Cases (TotalSegmentator dataset v2)}}
  \label{tab:cases}
  \scriptsize
  \setlength{\tabcolsep}{3pt}
  \begin{tabular}{clllrrrr}
    \toprule
    Case & ID & Description & Matrix (voxels) & Voxels (million) & MacBook Pro (s) & Windows desktop (s) & MacBook Air 2017 (s) \\
    \midrule
    1 & s0289 & Head CT angiography & 145 $\times$ 145 $\times$ 233 & 4.9 & 5.4 & 11.1 & 46 \\
    2 & s0138 & Heart and thoracic aorta & 221 $\times$ 221 $\times$ 215 & 10.5 & 12.7 & 25.4 & 82 \\
    3 & s1340 & Thorax to pelvis & 275 $\times$ 275 $\times$ 431 & 32.6 & 18.5 & 37.9 & 131 \\
    4 & s0970 & Polytrauma & 333 $\times$ 333 $\times$ 511 & 56.7 & 20.8 & 43.2 & 166 \\
    5 & s1121 & Pelvis-to-leg CT angiography & 281 $\times$ 281 $\times$ 519 & 41.0 & 20.5 & 42.4 & 148 \\
    6 & s0931 & Neck-to-leg CT angiography & 245 $\times$ 187 $\times$ 479 & 21.9 & 8.0 & 16.6 & 60 \\
    \bottomrule
  \end{tabular}

  \vspace{4pt}
  \begin{minipage}{\textwidth}
  \footnotesize All images are resampled to 1.5-mm isotropic voxels in the released dataset; the last dimension is the number of slices. The last three columns give the conversion time, one pass at the final cell size, for two million splats.
  \end{minipage}
\end{table}

\begin{table}[h]
  \centering
  \caption{\textbf{Paired comparisons of each variant with the full method (two million splats)}}
  \label{tab:paired}
  \footnotesize
  \setlength{\tabcolsep}{4pt}
  \begin{tabular}{lllrrrrrr}
    \toprule
    Look & Views & Variant & $\Delta$PSNR (dB) & $p$ & $\Delta$SSIM & $p$ & $\Delta$LPIPS & $p$ \\
    \midrule
    Bone & wide & 1 M splats & $-0.885$ & 0.156 & $-0.003$ & 0.312 & $+0.005$ & 0.094 \\
    Bone & wide & 3 M splats & $+0.036$ & 1.000 & $-0.000$ & 0.562 & $-0.001$ & 0.438 \\
    Bone & wide & No anti-aliasing compensation & $-3.832$ & 0.031 & $-0.010$ & 0.031 & $+0.001$ & 0.562 \\
    Bone & wide & No opacity mapping & $-4.498$ & 0.031 & $-0.012$ & 0.062 & $+0.001$ & 1.000 \\
    Bone & wide & Without faint splats & $-0.274$ & 0.031 & $-0.001$ & 0.031 & $+0.001$ & 0.031 \\
    Bone & wide & Uniform grid (1 cell/voxel) & $-0.171$ & 0.438 & $-0.000$ & 0.438 & $+0.001$ & 0.219 \\
    Bone & wide & Uniform grid vs octree, equal count & $+0.393$ & 0.031 & $+0.001$ & 0.031 & $-0.001$ & 0.031 \\
    Bone & close-up & 1 M splats & $-1.239$ & 0.031 & $-0.010$ & 0.031 & $+0.010$ & 0.031 \\
    Bone & close-up & 3 M splats & $+0.614$ & 0.031 & $+0.006$ & 0.031 & $-0.006$ & 0.062 \\
    Bone & close-up & No anti-aliasing compensation & $-0.398$ & 0.062 & $-0.002$ & 0.094 & $-0.001$ & 0.031 \\
    Bone & close-up & No opacity mapping & $-5.061$ & 0.031 & $-0.037$ & 0.031 & $+0.005$ & 0.562 \\
    Bone & close-up & Without faint splats & $-0.445$ & 0.062 & $-0.006$ & 0.062 & $-0.004$ & 0.031 \\
    Bone & close-up & Uniform grid (1 cell/voxel) & $-0.594$ & 0.062 & $-0.006$ & 0.062 & $+0.005$ & 0.062 \\
    Bone & close-up & Uniform grid vs octree, equal count & $+0.554$ & 0.031 & $+0.006$ & 0.031 & $-0.006$ & 0.031 \\
    \midrule
    Vessels & wide & 1 M splats & $-0.926$ & 0.031 & $-0.007$ & 0.031 & $+0.007$ & 0.031 \\
    Vessels & wide & 3 M splats & $+0.445$ & 0.062 & $+0.002$ & 0.156 & $-0.003$ & 0.062 \\
    Vessels & wide & No anti-aliasing compensation & $-0.854$ & 0.031 & $-0.005$ & 0.062 & $+0.001$ & 0.312 \\
    Vessels & wide & No opacity mapping & $-1.327$ & 0.031 & $-0.010$ & 0.031 & $+0.004$ & 0.062 \\
    Vessels & wide & Without faint splats & $-0.385$ & 0.031 & $-0.002$ & 0.031 & $+0.003$ & 0.031 \\
    Vessels & wide & Uniform grid (1 cell/voxel) & $-0.035$ & 0.562 & $+0.000$ & 1.000 & $+0.000$ & 0.562 \\
    Vessels & wide & Uniform grid vs octree, equal count & $-0.004$ & 0.438 & $+0.000$ & 0.042 & $+0.000$ & 0.438 \\
    Vessels & close-up & 1 M splats & $-1.313$ & 0.031 & $-0.016$ & 0.031 & $+0.024$ & 0.031 \\
    Vessels & close-up & 3 M splats & $+0.764$ & 0.031 & $+0.008$ & 0.031 & $-0.011$ & 0.031 \\
    Vessels & close-up & No anti-aliasing compensation & $-0.049$ & 0.094 & $-0.000$ & 0.312 & $-0.000$ & 0.844 \\
    Vessels & close-up & No opacity mapping & $-1.926$ & 0.031 & $-0.018$ & 0.031 & $+0.009$ & 0.031 \\
    Vessels & close-up & Without faint splats & $-0.528$ & 0.031 & $-0.005$ & 0.031 & $+0.001$ & 0.688 \\
    Vessels & close-up & Uniform grid (1 cell/voxel) & $-0.184$ & 0.562 & $-0.000$ & 1.000 & $+0.001$ & 0.562 \\
    Vessels & close-up & Uniform grid vs octree, equal count & $-0.055$ & 0.438 & $+0.000$ & 0.438 & $-0.000$ & 0.156 \\
    \midrule
    Skin and bone & wide & 1 M splats & $-0.645$ & 0.062 & $-0.008$ & 0.031 & $+0.011$ & 0.031 \\
    Skin and bone & wide & 3 M splats & $-0.115$ & 0.562 & $+0.002$ & 0.438 & $-0.004$ & 0.031 \\
    Skin and bone & wide & No anti-aliasing compensation & $-0.308$ & 0.156 & $+0.000$ & 1.000 & $-0.002$ & 0.031 \\
    Skin and bone & wide & No opacity mapping & $-5.274$ & 0.031 & $-0.015$ & 0.031 & $-0.001$ & 0.562 \\
    Skin and bone & wide & Without faint splats & $-1.632$ & 0.031 & $-0.003$ & 0.031 & $-0.001$ & 0.438 \\
    Skin and bone & wide & Uniform grid (1 cell/voxel) & $-0.242$ & 0.219 & $-0.003$ & 0.312 & $+0.006$ & 0.438 \\
    Skin and bone & wide & Uniform grid vs octree, equal count & $-0.125$ & 0.438 & $+0.000$ & 0.893 & $-0.001$ & 0.438 \\
    Skin and bone & close-up & 1 M splats & $-0.857$ & 0.031 & $-0.009$ & 0.031 & $+0.012$ & 0.031 \\
    Skin and bone & close-up & 3 M splats & $+0.355$ & 0.031 & $+0.004$ & 0.094 & $-0.007$ & 0.031 \\
    Skin and bone & close-up & No anti-aliasing compensation & $+0.112$ & 0.031 & $+0.000$ & 0.031 & $-0.000$ & 0.031 \\
    Skin and bone & close-up & No opacity mapping & $-5.866$ & 0.031 & $-0.025$ & 0.031 & $+0.001$ & 0.438 \\
    Skin and bone & close-up & Without faint splats & $-1.382$ & 0.031 & $-0.004$ & 0.062 & $-0.004$ & 0.094 \\
    Skin and bone & close-up & Uniform grid (1 cell/voxel) & $-0.521$ & 0.094 & $-0.006$ & 0.156 & $+0.007$ & 0.156 \\
    Skin and bone & close-up & Uniform grid vs octree, equal count & $-0.073$ & 0.562 & $+0.000$ & 0.844 & $-0.003$ & 0.062 \\
    \bottomrule
  \end{tabular}

  \vspace{4pt}
  \begin{minipage}{\textwidth}
  \footnotesize Median of the per-case differences (variant minus full method) over six cases; metrics were first averaged over the 48 wide or 16 close-up views of each scene. $p$: two-sided exact Wilcoxon signed-rank test over six cases (smallest attainable value 0.031); exploratory, not adjusted for multiple comparisons.
  \end{minipage}
\end{table}

\begin{table}[h]
  \centering
  \caption{\textbf{Thick slices: case 1, PSNR (dB) / SSIM / LPIPS}}
  \label{tab:thick}
  \small
  \begin{tabular}{llccc}
    \toprule
    Look & Views & 1.5-mm data, millimetre cells & 5-mm slices, millimetre cells & 5-mm slices, voxel-unit cells \\
    \midrule
    Bone & wide & 33.61 / 0.982 / 0.031 & 34.33 / 0.983 / 0.032 & 22.54 / 0.898 / 0.109 \\
    Bone & close-up & 30.03 / 0.961 / 0.178 & 31.67 / 0.966 / 0.193 & 17.63 / 0.723 / 0.367 \\
    Vessels & wide & 30.22 / 0.966 / 0.042 & 31.02 / 0.964 / 0.043 & 24.64 / 0.901 / 0.112 \\
    Vessels & close-up & 27.39 / 0.925 / 0.156 & 28.00 / 0.924 / 0.181 & 20.07 / 0.761 / 0.309 \\
    Skin and bone & wide & 32.26 / 0.968 / 0.088 & 32.87 / 0.970 / 0.087 & 20.46 / 0.843 / 0.171 \\
    Skin and bone & close-up & 29.36 / 0.942 / 0.306 & 29.77 / 0.950 / 0.289 & 16.34 / 0.756 / 0.413 \\
    \bottomrule
  \end{tabular}

  \vspace{4pt}
  \begin{minipage}{\textwidth}
  \footnotesize Two million splats in all three conditions. Each condition is compared with the reference rendered from the same data (1.5-mm or 5-mm).
  \end{minipage}
\end{table}

\end{document}